\documentclass[lettersize,journal]{IEEEtran}
\usepackage{amsmath,amsfonts}
\usepackage{algorithm}
\usepackage{array}
\usepackage[caption=false,font=normalsize,labelfont=sf,textfont=sf]{subfig}
\usepackage{textcomp}
\usepackage{stfloats}
\usepackage{url}
\usepackage{verbatim}
\usepackage{graphicx}
\usepackage{cite}

\usepackage{bm}
\usepackage{algpseudocode}
\usepackage{diagbox}
\usepackage{multicol}
\usepackage{times}

\usepackage{soul}
\usepackage[utf8]{inputenc}
\usepackage{makecell}
\usepackage{float}
\usepackage{stfloats}
\usepackage[OT1]{fontenc}

\usepackage{amssymb}
\usepackage{amsthm}
\usepackage{mathrsfs}
\usepackage{stmaryrd}
\usepackage{graphicx,amsfonts,listings,xcolor,colortbl}

\usepackage{booktabs}
\usepackage{tabularx}
\usepackage{blkarray}
\usepackage{arydshln}
\usepackage{fancyhdr}
\usepackage{setspace}
\usepackage{multirow}
\usepackage{color}
\usepackage{xcolor}

\usepackage{threeparttable}

\begin{document}

\title{Structure-aware Hybrid-order Similarity Learning for Multi-view Unsupervised Feature Selection}

\author{Lin~Xu,~Ke~Li,~Dongjie Wang,~Fengmao~Lv,~Tianrui Li,~\IEEEmembership{Senior Member,~IEEE}, and~Yanyong~Huang
    \thanks{Lin~Xu, Ke~Li, and Yanyong~Huang are with the Joint Laboratory of Data Science and Business Intelligence, School of Statistics and Data Science, Southwestern University of Finance and Economics, Chengdu 611130, China, and also with the Big Data Laboratory on Financial Security and Behavior, SWUFE (Laboratory of Philosophy and Social Sciences, Ministry of Education), Chengdu 611130, China (e-mail: xulin00@foxmail.com; likec@swufe.edu.cn; huangyy@swufe.edu.cn), Yanyong Huang is the corresponding author;}
    \thanks{Dongjie Wang is with the Department of Electrical Engineering and Computer Science, University of Kansas, Lawrence, KS 66045, USA (e-mail: wangdongjie@ku.edu);}
    \thanks{Fengmao~Lv  and Tianrui Li are with the School of Computing and Artificial Intelligence, Southwest Jiaotong University, Chengdu 611756, China (e-mail: fengmaolv@126.com; trli@swjtu.edu.cn).}
}
\markboth{Journal of \LaTeX\ Class Files,~Vol.~14, No.~8, August~2021}%
{Shell \MakeLowercase{\textit{et al.}}: A Sample Article Using IEEEtran.cls for IEEE Journals}


\maketitle

\begin{abstract}
Multi-view unsupervised feature selection (MUFS) has recently emerged as an effective dimensionality reduction method for unlabeled multi-view data. However, most existing methods mainly use first-order similarity graphs to preserve local structure, often overlooking the global structure that can be captured by second-order similarity. In addition, a few MUFS methods leverage predefined second-order similarity graphs, making them vulnerable to noise and outliers and resulting in suboptimal feature selection performance. In this paper, we propose a novel MUFS method, termed Structure-aware Hybrid-order sImilarity learNing for multi-viEw unsupervised Feature Selection (SHINE-FS), to address the aforementioned problem. SHINE-FS first learns consensus anchors and the corresponding anchor graph to capture the cross-view relationships between the anchors and the samples.  Based on the acquired cross-view consensus information, it generates low-dimensional representations of the samples, which facilitate the reconstruction of multi-view data by identifying discriminative features. Subsequently, it employs the anchor-sample relationships to learn a second-order similarity graph. Furthermore, by jointly learning first-order and second-order similarity graphs, SHINE-FS constructs a hybrid-order similarity graph that captures both local and global structures, thereby revealing the intrinsic data structure to enhance feature selection. Comprehensive experimental results on real multi-view datasets show that SHINE-FS outperforms the state-of-the-art methods.
\end{abstract}


\section{Introduction}
\IEEEPARstart{W}{ith} the rapid advancement of information technology, multi-view data has become increasingly common in real-world applications, where each sample is described by heterogeneous features from different perspectives~\cite{li2017feature, pan2023low}. However, the high dimensionality of multi-view data often results in the curse of dimensionality, significantly reducing the performance of downstream tasks~\cite{zhang2019feature, qiang2025adaptive}. Moreover, labeling such data is difficult due to the high cost of manual annotation~\cite{hou2017multi, jiang2022semi}. To address these challenges, multi-view unsupervised feature selection (MUFS) effectively reduces the dimensionality of unlabeled multi-view data by selecting a compact subset of informative features without relying on labels~\cite{yang2024self,SCMvFS2023}.

Existing MUFS methods can be broadly divided into two categories. The first concatenates features from all views and then applies single-view feature selection methods, such as LPScore~\cite{LS2005}, HSL~\cite{HSL2024}, and GAWFS~\cite{GAWFS2025}. In contrast, the second category selects discriminative features directly from multi-view data by leveraging inter-view correlations. Representative methods include JMVFG~\cite{JMVFG2024}, GCDUFS~\cite{GCDUFS2025}, and UKMFS~\cite{UKMFS2025}. Leveraging similarity graphs to preserve local structure has been shown to significantly improve MUFS performance, leading to the development of numerous graph-based approaches. JMVFG~\cite{JMVFG2024} employs orthogonal decomposition to obtain both view-specific basis matrices and a shared clustering indicator matrix, while simultaneously learning a consensus similarity graph to guide feature selection. GCDUFS~\cite{GCDUFS2025} learns a consensus graph to capture cross-view information and derives diverse representations from different views to exploit their complementary information, both of which help identify discriminative features. UKMFS~\cite{UKMFS2025} unifies multi-view representations via nonlinear kernel mapping, learns a consistent graph under low-rank constraints, and utilizes binary hashing codes to facilitate feature selection. 

\begin{figure*}[t]
    \centering
    \includegraphics[width=1.956\columnwidth]{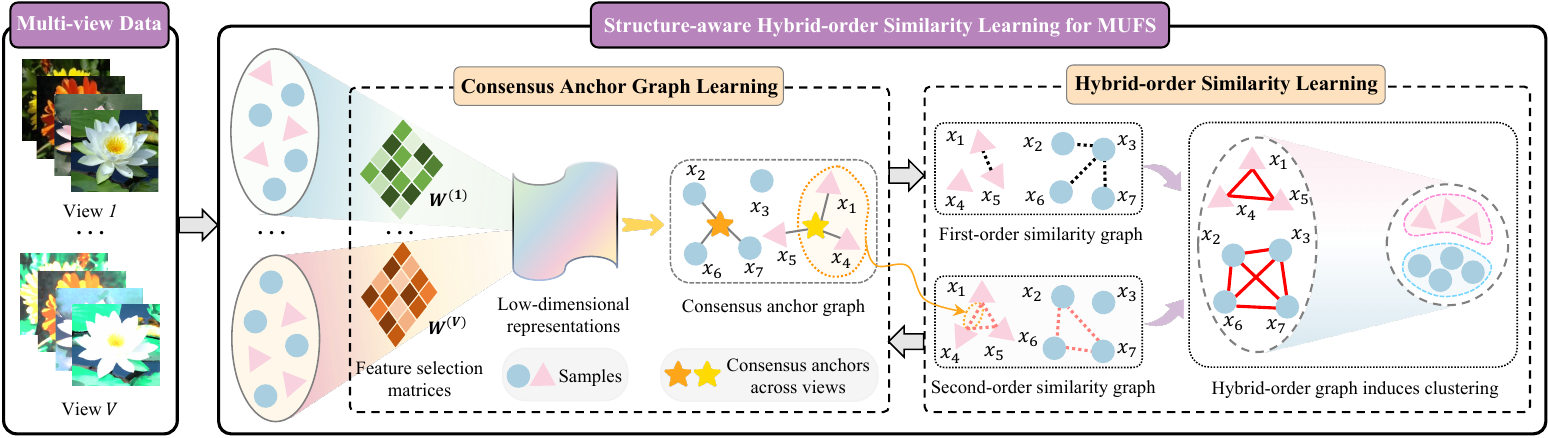} 
\caption{The framework of the proposed SHINE-FS.}
\label{framework}
\end{figure*}

Although existing graph-based methods have achieved promising performance, they typically rely on first-order similarity graphs to preserve local data structure~\cite{zhang2024scalable}, while often neglecting global structures captured by second-order similarity graphs~\cite{LINE2015}. For example, in document clustering, a first-order similarity graph connects only documents that are directly similar in content, thus making them immediate neighbors. Extending beyond direct pairwise similarities, a second-order similarity graph captures topic-level groupings by linking documents that share many common neighbors—even if they are not directly similar—since these documents are connected to the same set of other documents. This enables the second-order graph to capture more global relationships within the corpus~\cite{thijs2013second}. A few studies have investigated unsupervised feature selection for second-order graphs. HSL~\cite{HSL2024} exploits first-order and higher-order similarities in the low-dimensional embedding space to capture structural information, and learns an optimal graph Laplacian for selecting discriminative features. CFSMO~\cite{CFSMO2024} projects multi-view data into a shared latent space to capture complementary information, and integrates multi-order structural information across multiple views to guide feature selection. These approaches typically predefine second-order graphs before performing feature selection. However, their vulnerability to noise and outliers often compromises feature selection performance.

To address the aforementioned issues, we propose a novel MUFS method called Structure-aware Hybrid-order sImilarity learNing for multi-viEw unsupervised Feature Selection (SHINE-FS). Specifically, we first learn a set of consensus anchors and the corresponding anchor graph to capture cross-view relationships between anchors and samples. Next, we derive low-dimensional representations of the samples from the learned consensus information, enabling the reconstruction of multi-view data and the identification of discriminative features. Subsequently, we learn a second-order similarity graph based on anchor-sample relationships within the consensus anchor graph, where samples strongly connected to the same anchor are considered to share a second-order relationship. Furthermore, by jointly learning first-order and second-order similarity graphs to construct a hybrid-order graph, SHINE-FS effectively captures both local and global structural information, thereby better preserving the intrinsic data structure and guiding feature selection. Finally, an iterative optimization algorithm is developed for the proposed model, and extensive experiments on multiple real-world multi-view datasets are conducted to demonstrate the effectiveness of SHINE-FS over state-of-the-art methods. The framework of SHINE-FS is illustrated in Fig.~\ref{framework}. 

The main contributions of this work can be summarized as follows:
(\romannumeral1) Instead of limiting MUFS to the use of a first-order similarity graph or simply combining it with a predefined second-order graph, we propose a hybrid-order similarity graph adaptive learning approach that incorporates consensus anchor graph-induced regularization, which fully leverages both local and global structural information from graphs of different orders to enhance feature selection. (\romannumeral2) SHINE-FS seamlessly integrates multi-view feature selection, consensus anchor graph learning, and hybrid-order similarity learning into a unified framework, allowing these components to mutually reinforce each other and ultimately improve feature selection performance.
(\romannumeral3) We develop an effective and fast-converging iterative optimization algorithm to implement the proposed method. Extensive experimental results on real-world multi-view datasets demonstrate that SHINE-FS outperforms state-of-the-art approaches.

The remainder of this paper is organized as follows. Section II reviews related work on MUFS. In Section III, the proposed SHINE-FS method is presented, together with the iterative optimization algorithm developed to address the formulated optimization problem, as well as analyses of the algorithm's time complexity and convergence. Extensive experimental results on real multi-view datasets are reported in Section IV. Section V concludes the paper.

\section{Related Work}

In this section, we briefly introduce some representative MUFS methods, which can be broadly categorized into single-view and multi-view unsupervised feature selection approaches.

To select informative features from multi-view data, the first category of MUFS approaches concatenates features from all views and subsequently conducts feature selection using conventional single-view methods, typically including LPScore~\cite{LS2005} and GAWFS~\cite{GAWFS2025}. LPScore identifies informative features by using the Laplacian score, which evaluates how well each feature preserves the local structure of the data~\cite{LS2005}. GAWFS combines non-negative matrix factorization with adaptive graph learning to discover discriminative features through a learned feature weighting matrix~\cite{GAWFS2025}. Jiang et al.~\cite{jiang2025block} developed a graph-based unsupervised feature selection approach that constructs a block-diagonal graph within a low-dimensional subspace, thereby effectively capturing the intrinsic data structure to inform the feature selection process. The single-view-based methods mentioned above address multi-view data by simply merging features from different views, but they overlook correlations between views, which can result in suboptimal feature selection. 

In contrast, the second category of MUFS methods directly selects discriminative features from multi-view data by leveraging inter-view correlations to enhance performance. Zhou et al.~\cite{SDFS2023} introduced a structural regularization-based MUFS method that preserves data locality by jointly learning both view-specific and consensus similarity graphs throughout the feature selection process. JMVFG~\cite{JMVFG2024} maintains cross-view consistency by decomposing the projected data matrix into view-specific basis matrices and a clustering indicator matrix, while simultaneously learning a consensus graph that preserves the local data structure to facilitate feature selection. Yuan et al.~\cite{TLR2022} promoted cross-view consistency through tensor low-rank regularization and utilized intra-view structural information from view-specific graphs for feature selection. GCDUFS~\cite{GCDUFS2025} captures consensus information across views via graph regularization and explores inter-view diversity using kernel-based dependency measures, integrating both aspects into the feature selection process. Zhang et al.~\cite{EMUFS2024} developed an efficient MUFS method that collaboratively fuses membership matrices and bipartite graphs to capture both cluster structure and local data structure for discriminative feature selection. UKMFS~\cite{UKMFS2025} unifies multi-view data representations through kernel mapping, learns a consistent graph across views using low-rank constraints, and employs binary hashing codes to guide feature selection. Cao et al.~\cite{CDMvFS2024} utilized view-specific similarity graphs to explore a consensus clustering structure, aiding feature selection by leveraging both intra-view and inter-view information. CCSFS~\cite{CCSFS2023} identifies discriminative features by learning a consensus cluster indicator matrix, which is obtained by fusing partition matrices from multiple views.

Although these methods achieve promising results in feature selection, they typically preserve the local data structure by utilizing first-order similarity graphs, while often overlooking the global structures captured by second-order graphs. Only a few studies have explored the use of second-order graphs for feature selection. HSL~\cite{HSL2024} projects high-dimensional data into a low-dimensional space, preserving both first-order and higher-order similarity structures to facilitate the identification of discriminative features. CFSMO~\cite{CFSMO2024} selects features by integrating neighbor information from multi-order similarity graphs with consistency information from a shared latent representation. The two methods mentioned above typically predefine second-order graphs, making them vulnerable to noise and outliers and ultimately diminishing the effectiveness of feature selection.

\section{The Proposed Method}

In this section, we begin by summarizing the notations and definitions used throughout the paper. We then introduce our proposed method SHINE-FS, and develop an iterative algorithm to solve the associated optimization problem. Finally, we analyze the time complexity and convergence of the algorithm.

\subsection{Notations and Definitions}
Throughout this paper, matrices are represented by boldface uppercase letters. For a matrix $\bm{M} \in \mathbb{R}^{p \times q}$, $\bm{M}_{i \cdot}$ and $\bm{M}_{\cdot j}$ denote its $i$-th row and $j$-th column, respectively, and $\mathnormal{M}_{ij}$ represents the $(i,j)$-th entry. The transpose of $\bm{M}$ is denoted by $\bm{M}^T$, and its trace is $\operatorname{Tr}(\bm{M})$. The Frobenius norm of $\bm{M}$ is defined as $\|\bm{M}\|_{\mathrm{F}}=\sqrt{\sum_{i=1}^{p}\sum_{j=1}^{q}\mathnormal{M}_{ij}^{2}}$, and the $\ell_{2,1}$-norm is given by $\|\bm{M}\|_{2,1}=\sum_{i=1}^{p}\sqrt{\sum_{j=1}^{q}\mathnormal{M}_{ij}^{2}}$. Let $\|\bm{M}\|_0$ denote the $\ell_0$-norm of $\bm{M}$, which counts the number of non-zero elements in $\bm{M}$. $\bm{I}$ denotes the identity matrix. $\bm{1}_n = [1, \dots, 1]^T$ is an $n$-dimensional column vector of ones. It is denoted by $\bm{1}$ when the size is clear from the context. Let $\mathcal{X} = \{\bm{X}^{(v)}\}_{v=1}^l$ represent a multi-view dataset with $l$ views, where $\bm{X}^{(v)} \in \mathbb{R}^{d_v \times n}$ denotes the data matrix for the $v$-th view, consisting of $n$ instances and $d_v$ features. Our goal is to select the $h$ most discriminative features from the multi-view dataset $\mathcal{X}$.

\subsection{Formulation of SHINE-FS}

Previous studies have shown that consensus information among multiple views is essential for preserving their shared structure, which in turn facilitates feature selection~\cite{GCDUFS2025,CEUMFS2024,TRCA2023}. Existing anchor-based MUFS methods typically select representative samples as anchors to construct view-specific anchor graphs, which are then fused to reveal the consensus structure~\cite{EMUFS2024,zhang2024scalable}. However, noise in the original feature space can prevent these predefined anchors from accurately capturing the underlying data structure. Additionally, these methods do not leverage anchor–sample relationships across views to enable second-order graph learning, which is vital for uncovering the global structure of the data~\cite{LINE2015}. To address these limitations, we propose an adaptive anchor-induced second-order graph learning approach to guide multi-view feature selection, as described below:
\begin{equation}\label{fun-anchor}
	\begin{aligned}
		\min _{\substack{\alpha^{(v)},\bm{W}^{(v)},\\ \bm{C},\bm{A},\bm{S}}} &\sum_{v=1}^{l}(\alpha^{(v)})^{2} [\|{\bm{X}}^{(v)}-\bm{W}^{(v)}\bm{C}\bm{A}\|_{F}^{2} + \gamma\|\bm{W}^{(v)}\|_{2,1}]\\
		&+ \sum_{i,j=1}^{n} \|\bm{A}_{\cdot i} - \bm{A}_{\cdot j}\|_{2}^{2}\mathnormal{S}_{ij} + \beta\|\bm{A}\|_{F}^{2}+\lambda_{S}\|\bm{S}\|_{F}^{2}\\
		\text { s.t. }~ & 0 \!\leq\! \alpha^{(v)} \!\leq\! 1,\!\sum_{v=1}^{l}\!\alpha^{(v)}\!=\!1,\! {\bm{W}^{(v)}}^T \!\bm{W}^{(v)}\!=\!\!\bm{I},\!\bm{C}^T\!\bm{C}\!=\!\!\bm{I},\\
        &{\bm{A}}^T {\bm{1}} = {\bm{1}},{\mathnormal{A}}_{ij} \ge 0,{\| {{\bm{S}}_{i \cdot }} \|_0} = k,{\bm{S}}_{i \cdot }{\bm{1}} = 1,{\mathnormal{S}}_{ij} \ge 0,\\	
	\end{aligned}
\end{equation}
where $\alpha^{(v)}$ denotes the weight of the $v$-th view, $\bm{C} \in \mathbb{R}^{c \times m}$ represents a set of $m$ consensus anchors, each being a $c$-dimensional vector derived from the unified latent space, $\bm{A} \in \mathbb{R}^{m \times n}$ is the consensus anchor graph, and $\bm{S} \in \mathbb{R}^{n \times n}$ represents the anchor-induced second-order similarity graph. By leveraging the direct anchor–sample relationships across views in $\bm{A}$, $\bm{S}$ captures indirect second-order relationships by linking samples that share common anchors, in accordance with the second-order graph principle whereby a neighbor’s neighbor is also considered a neighbor~\cite{wang2023multi}. Additionally, $\bm{C}\bm{A}$ yields low-dimensional representations of the samples, wherein each sample is characterized by its associations to the consensus anchors. The matrix $\bm{W}^{(v)} \in \mathbb{R}^{d_v \times c}$  facilitates the reconstruction of the original data in the $v$-th view from its corresponding  low-dimensional representations, thereby quantifying the contribution of each feature to the reconstruction process. Furthermore,  applying the  $\ell_{2,1}$-norm regularization on $\bm{W}^{(v)}$ promotes row-wise sparsity, helping to suppress redundant or irrelevant features while retaining the important ones.

In Eq.~\eqref{fun-anchor}, rather than relying on predefined anchors, we adaptively learn a set of consensus anchors $\bm{C}$ and the corresponding anchor graph $\bm{A}$, which together capture anchor–sample relationships across multiple views. From the learned consensus information, we derive low-dimensional representations of the samples. These representations are then used to reconstruct the original multi-view data by identifying discriminative features. In addition, samples that share similar anchor neighbors can preserve their second-order relationships, thereby enabling effective characterization of the global data structure.

To leverage both local and global structural information for feature selection, we propose an adaptive hybrid-order similarity learning method that combines first-order and second-order similarity graphs,  as detailed below:
\begin{equation}\label{hybrid order}
	\begin{aligned}
		\min _{\substack{\bm{G},\bm{F}, \\ \alpha^{(v)}}} & \sum_{v=1}^{l}(\alpha^{(v)})^{2} \operatorname{Tr}(\bm{X}^{(v)} \bm{L}_{\bm{G}} {\bm{X}^{(v)}}^T) + \lambda_{G}\|\bm{G}\|_{F}^{2} \\
		&+ \frac{1}{2}\sum\limits_{i,j = 1}^n \| \bm{F}_{i \cdot } - \bm{F}_{j \cdot } \|_2^2 (\bm{G}+\eta\bm{S})_{ij}\\
		\text { s.t. }~ & 0 \leq \alpha^{(v)} \leq 1,\sum_{v=1}^{l} \alpha^{(v)} = 1, {\| {{\bm{G}}_{i \cdot }} \|_0} = k,{\bm{G}}_{i \cdot } {\bm{1}} = 1,\\
        & \mathnormal{G}_{ij} \ge 0, \bm{F}^T \bm{F}=\bm{I},\\	
	\end{aligned}
\end{equation}
where $\bm{G}$ denotes the first-order similarity graph, $\bm{L}_{\bm{G}}$ is the corresponding  Laplacian matrix, $\bm{G} + \eta \bm{S}$ represents the hybrid-order similarity graph, and $\bm{F}$  corresponds to the cluster indicator matrix. The Frobenius norm term  $\|\bm{G}\|_{F}^{2}$ is included to prevent trivial solutions, with $\lambda_{G}$ 
serving as its regularization parameter. Additionally, $\eta$  acts as a trade-off parameter, balancing the contributions of first-order and second-order similarities. In Eq.~\eqref{hybrid order}, the first-order similarity  graph $\lambda_{G}$ is adaptively learned to capture direct relationships among samples, thereby preserving the local structure of the data. Meanwhile, the anchor-induced second-order similarity graph $\bm{S}$ utilizes shared anchor neighborhood information from the consensus anchor graph to identify second-order indirect correlations between samples, capturing the global data structure. These two graphs are then integrated into a hybrid-order graph, which jointly leverages both local and global structural information to better characterize the intrinsic structure of the data. The resulting hybrid-order graph further guides clustering, ensuring that highly similar samples are assigned similar pseudo-labels.

By combining Eqs.~\eqref{fun-anchor} and~\eqref{hybrid order}, the overall objective function of SHINE-FS is summarized as follows:
\begin{equation}\label{fun-all}
	\begin{aligned}
		\min _{\bm{\Omega}} &\sum_{v=1}^{l}(\alpha^{(v)})^{2}[\|\bm{X}^{(v)}-\bm{W}^{(v)}\bm{C}\bm{A}\|_{F}^{2}\!+\!\operatorname{Tr}(\bm{X}^{(v)} \bm{L}_{\bm{G}} {\bm{X}^{(v)}}^T)\\
        &+\!\! \gamma\|\bm{W}^{(v)}\|_{2,1}\!] \!+\!\!\! \sum_{i,j=1}^{n}\!\!\!\|\bm{A}_{\cdot i} \!-\! \bm{A}_{\cdot j}\|_{2}^{2}\mathnormal{S}_{ij}\!+\!\beta\|\bm{A}\|_{F}^{2} \!+\!\! \lambda_{S}\|\bm{S}\|_{F}^{2} \\
		&+ \frac{1}{2}\sum\limits_{i,j = 1}^n \| \bm{F}_{i \cdot } - \bm{F}_{j \cdot } \|_2^2 (\bm{G}+\eta\bm{S})_{ij} + \lambda_{G}\|\bm{G}\|_{F}^{2}\\
		\text { s.t. }~ & 0 \leq \alpha^{(v)} \leq 1,\sum_{v=1}^{l} \alpha^{(v)}=1,{{\bm{A}}}^T {\bm{1}} = {\bm{1}},{\mathnormal{A}}_{ij} \ge 0,\\
        &{\| {{\bm{G}}_{i \cdot }} \|_0} \!=\! k,{\bm{G}}_{i \cdot } {\bm{1}} = 1,{\mathnormal{G}}_{ij} \ge 0,{\bm{W}^{(v)}}^T{\bm{W}^{(v)}}=\bm{I},\\
		&{\| {{\bm{S}}_{i \cdot }} \|_0} \!=\! k,{\bm{S}}_{i \cdot }{\bm{1}} \!=\! 1,{\mathnormal{S}}_{ij} \ge 0,\bm{F}^T\bm{F}\!=\!\bm{I},\bm{C}^T \! \bm{C}\!=\!\bm{I},\\	
	\end{aligned}
\end{equation}
where $\bm{\Omega}=\{\bm{C},\bm{A},\bm{G},\bm{S},\bm{F},\{\bm{W}^{(v)},\alpha^{(v)}\}_{v=1}^{l}\}$. Here, $\gamma$, $\beta$, and $\eta$ are hyperparameters, whereas $\lambda_{S}$ and $\lambda_{G}$ are automatically determined during optimization of $\bm{S}$ and $\bm{G}$, respectively.

Eq.~\eqref{fun-all} integrates multi-view feature selection, consensus anchor graph learning, and hybrid-order similarity learning into a unified framework. This unified framework leverages both local and global structural information to accurately characterize the intrinsic structure, thereby facilitating feature selection. Moreover, selecting discriminative features enhances the representativeness of the consensus anchors and the quality of the corresponding anchor graph, which in turn promotes the adaptive learning of hybrid-order similarity. These components mutually reinforce each other, thereby improving feature selection.

\subsection{Optimization and Algorithm}
As the objective function in Eq.~\eqref{fun-all} is not jointly convex with respect to all variables, we propose an iterative optimization algorithm that alternately updates each variable while keeping the others fixed.

\subsubsection{Updating $\bm{W}^{(v)}$ by Fixing Other Variables} 
When the other variables are fixed, the optimization subproblem w.r.t. $\bm{W}^{(v)}$ can be rewritten in the following equivalent trace form:
\begin{equation}\label{op-w2}
	\begin{aligned}
		\min_{{\bm{W}^{(v)}}} &\!\operatorname{Tr}(\gamma{\bm{W}^{(v)}}^T{\bm{D}^{(v)}}{\bm{W}^{(v)}}\!-\!2{\bm{W}^{(v)}}^T\bm{X}^{(v)}\bm{A}^T\bm{C}^T)\\
		\text { s.t. }~ & {\bm{W}^{(v)}}^T{\bm{W}^{(v)}}=\bm{I},	
	\end{aligned}
\end{equation}
where $\bm{D}^{(v)}$ denotes a diagonal matrix with the $i$-th diagonal element defined as $\bm{D}^{(v)}_{ii} = \frac{1}{2\sqrt{\| \bm{W}_{i \cdot }^{(v)} \|_2^2 + \varepsilon}}$ ($\varepsilon$ is a small constant added to avoid division by zero). Eq.~(\ref{op-w2}) can be efficiently solved using the generalized power iteration (GPI) method~\cite{nie2017generalized}.

\subsubsection{Updating $\bm{A}$ by Fixing Other Variables} 
With other variables fixed, the subproblem of solving $\bm{A}$ reduces to the following equivalent trace form:
\begin{equation}\label{op-a1}
	\begin{aligned}
		\min _{{{\bm{A}}}^T {\bm{1}} = {\bm{1}},{\mathnormal{A}}_{ij} \ge 0} &\operatorname{Tr}(\bm{P}{\bm{A}^T}\bm{A}-\bm{Q}\bm{A}),	
	\end{aligned}
\end{equation}
where $\bm{P} = [\sum_{v=1}^{l}(\alpha^{(v)})^{2}+\beta]\bm{I}+2\bm{L}_{\bm{S}}$, and $\bm{Q} = \sum_{v=1}^{l}2(\alpha^{(v)})^{2}{\bm{X}^{(v)}}^T{\bm{W}^{(v)}}\bm{C}$. According to~\cite{EMUFS2024}, the constraints on $\bm{A}$ are first relaxed to obtain a latent solution $\widehat{\bm{A}}$ by setting the derivative of Eq.~\eqref{op-a1} to zero. Then, the latent solution is projected onto the constrained space as follows:
\begin{equation}\label{op-a2}
	\begin{aligned}
		&\min _{{{\bm{A}}}^T {\bm{1}} = {\bm{1}},{\mathnormal{A}}_{ij} \ge 0}\|\bm{A}-\widehat{\bm{A}}\|_{F}^{2}.
	\end{aligned}
\end{equation}

According to~\cite{solution19}, the solution to Eq.~\eqref{op-a2} is given by:
\begin{equation}\label{op-a3}
	\begin{aligned}
		\bm{A} = (\widehat{\bm{A}}-\widehat{\bm{A}}\bm{1}_n\bm{1}_n^T+\bm{1}_m\bm{1}_n^T,0)_{+}.
	\end{aligned}
\end{equation}

\subsubsection{Updating $\bm{C}$ by Fixing Other Variables} 
By keeping the other variables fixed, solving $\bm{C}$ is equivalent to the following problem:
\begin{equation}\label{op-c1}
	\begin{aligned}
		\min _{\bm{C}^T\bm{C}=\bm{I}} &\sum_{v=1}^{l}(\alpha^{(v)})^{2}\|\bm{X}^{(v)}-\bm{W}^{(v)}\bm{C}\bm{A}\|_{F}^{2}.	
	\end{aligned}
\end{equation}

Then, Eq.~\eqref{op-c1} can be transformed into an equivalent trace form:
\begin{equation}\label{op-c2}
	\begin{aligned}
		\max _{\bm{C}^T\bm{C}=\bm{I}} &\operatorname{Tr}({\bm{C}}^T\bm{E}),\\
	\end{aligned}
\end{equation}
where $\bm{E} = \sum_{v=1}^{l} 2 (\alpha^{(v)})^{2}{\bm{W}^{(v)}}^T\bm{X}^{(v)}{\bm{A}}^T$. The optimal solution to Eq.~\eqref{op-c2} can be obtained via the singular value decomposition (SVD) of $\bm{E}$ as the product of its left and right singular vectors~\cite{wang2019multi}.

\subsubsection{Updating $\bm{S}$ by Fixing Other Variables} 
Since the rows of $\bm{S}$ are independent, $\bm{S}_{i \cdot}$ can be updated by solving the following equivalent problem after fixing the other variables:
\begin{equation}\label{op-s2}
\begin{aligned}
\mathop {\min }\limits_{{\bm{S}}_{i \cdot }  } \;& \frac{1}{2}\| {{\bm{S}}_{i \cdot }   + \frac{{{\bm{U}}_{i \cdot }  }}{{2{\lambda_{S}}}}} \|_2^2\\
\text { s.t. }~&{\| {{\bm{S}}_{i \cdot }} \|_0} = k,{\bm{S}}_{i \cdot } {\bm{1}} = 1,{\mathnormal{S}}_{ij} \ge 0,
\end{aligned}
\end{equation}
where $\bm{U}_{i \cdot}$ denotes a vector, with its $j$-th entry $\mathnormal{U}_{ij} = \sum_{r=1}^{m} (\mathnormal{A}_{ri} - \mathnormal{A}_{rj})^{2}+\frac{\eta}{2}\|\bm{F}_{i \cdot } - \bm{F}_{j \cdot }\|_{2}^{2}$. Following~\cite{nie2016unsupervised}, the solution for $\bm{S}$ is given by:
\begin{equation}\label{op-s3}
    \mathnormal{S}_{ij}  = \left\{ 
        \begin{aligned}
            &\frac{\mathnormal{U}_{i,k + 1}  - \mathnormal{U}_{ij} }{{k \mathnormal{U}_{i,k + 1} } - \sum_{p=1}^{k} \mathnormal{U}_{ip} }, j \le k; \\
            & \quad \quad \quad \;\;\;0\quad \quad \quad \quad \;\;,j > k.\\
        \end{aligned}
    \right.
\end{equation}

In addition, $\lambda_{S}$ is adaptively determined as $(k \mathnormal{U}_{i, k+1} - \sum_{p=1}^{k} \mathnormal{U}_{i p}) / 2$ to ensure that $\bm{S}_{i \cdot}$ contains exactly $k$ nonzero elements.

\subsubsection{Updating $\bm{G}$ by Fixing Other Variables} 
Following a similar procedure as in Eq.~\eqref{op-s2}, each row of $\bm{G}$ is updated by solving the following equivalent optimization problem with the other variables fixed:
\begin{equation}\label{op-g1}
	\begin{aligned}
        \mathop {\min }\limits_{{\bm{G}}_{i \cdot }  } \;& \frac{1}{2}\| {{\bm{G}}_{i \cdot }   + \frac{{{\bm{B}}_{i \cdot }  }}{{2{\lambda_{G}}}}} \|_2^2\\
		\text { s.t. }~&{\| {{\bm{G}}_{i \cdot }} \|_0} = k,{\bm{G}}_{i \cdot } {\bm{1}} = 1,{\mathnormal{G}}_{ij} \ge 0,		
	\end{aligned}
\end{equation}
where $\bm{B}_{i \cdot}$ is a vector with the $j$-th element $\mathnormal{B}_{ij} = \frac{1}{2}[\sum_{v=1}^{l} (\alpha^{(v)})^{2} \|\bm{X}^{(v)}_{\cdot i} - \bm{X}^{(v)}_{\cdot j}\|_{2}^{2}+\|\bm{F}_{i \cdot } - \bm{F}_{j \cdot }\|_{2}^{2}]$. Similar to solving Eq.~(\ref{op-s2}), the optimal solution for $\bm{G}$ can be obtained as follows:
\begin{equation}\label{op-g2}
    \mathnormal{G}_{ij}  = \left\{ 
        \begin{aligned}
            &\frac{\mathnormal{B}_{i,k + 1}  - \mathnormal{B}_{ij} }{{k \mathnormal{B}_{i,k + 1} } - \sum_{p=1}^{k} \mathnormal{B}_{ip} } , j \le k; \\
            & \quad \quad \quad \;\;\;\;\;0\quad \quad \quad \; \;\; \;,j > k.\\
        \end{aligned}
    \right.
\end{equation}

Meanwhile, $\lambda_{G}$ is set to $(k \mathnormal{B}_{i, k+1} - \sum_{p=1}^{k} \mathnormal{B}_{i p})/2$ to guarantee that $\bm{G}_{i \cdot }$ contains $k$ nonzero elements.

\subsubsection{Updating $\bm{F}$ by Fixing Other Variables} 
With all other variables fixed, the optimization subproblem for $\bm{F}$ can be formulated as follows:
\begin{equation}\label{op-f}
    \begin{aligned}
    \mathop {\min }\limits_{{{\bm{F}}^T}{\bm{F}} = {\bm{I}}} Tr( {{{\bm{F}}^T}{{\bm{L}}^{*}}{\bm{F}}} ),\\
    \end{aligned}
\end{equation}
where $\bm{L}^{*}$ denotes the Laplacian matrix associated with the hybrid-order similarity graph. The solution to $\bm{F}$ of Eq.~(\ref{op-f}) is given by the eigenvectors corresponding to the smallest $c$ eigenvalues of $\bm{L}^{*}$~\cite{nie2016unsupervised}.

\begin{algorithm}[t]
	\caption{Iterative algorithm of SHINE-FS}
	\label{code-all}
	\textbf{Input}: Multi-view data $\mathcal{X} = \{\bm{X}^{(v)}\}_{v=1}^{l}$, parameters $\gamma$, $\beta$, and $\eta$.
	\begin{algorithmic}[1]
		\State Initialize: $\{\bm{D}^{(v)}, \alpha^{(v)}\}_{v=1}^{l}$, $\bm{A}$, $\bm{C}$, $\bm{S}$, and $\bm{G}$.
		\While{not convergent}
		\State Update $\{\bm{W}^{(v)}\}_{v=1}^{l}$ by solving Eq.~\eqref{op-w2}.
		\State Update $\bm{A}$ according to Eq.~\eqref{op-a3}.
		\State Update $\bm{C}$ by solving Eq.~\eqref{op-c2}.
		\State Update $\bm{S}$ according to Eq.~\eqref{op-s3}.
		\State Update $\bm{G}$ according to Eq.~\eqref{op-g2}.
		\State Update $\bm{F}$ by solving Eq.~\eqref{op-f}.
		\State Update $\{\alpha^{(v)}\}_{v=1}^{l}$ according to Eq.~\eqref{op-alpha2}.
		\EndWhile
	\end{algorithmic} 
	\textbf{Output}: Sort the $\ell_{2}$-norms of the rows in $\{\bm{W}^{(v)}\}_{v=1}^{l}$ in descending order, and select the top $h$ features from $\mathcal{X}$.
\end{algorithm}

\subsubsection{Updating $\alpha^{(v)}$ by Fixing Other Variables}  
When fixing other variables, we update $\alpha^{(v)}$ by solving the following problem:
\begin{equation}\label{op-alpha1}
    \begin{aligned}
    \mathop {\min }\limits_{ \alpha^{(v)}} & \sum\limits_{v = 1}^l {(\alpha ^{(v)})^2 z^{(v)}} \\
    \text { s.t. }~ & 0 \leq \alpha^{(v)} \leq 1,\sum_{v=1}^{l} \alpha^{(v)}=1,
    \end{aligned}
\end{equation}
where $z^{(v)} = \|\bm{X}^{(v)}-\bm{W}^{(v)}\bm{C}\bm{A}\|_{F}^{2}+\gamma\|\bm{W}^{(v)}\|_{2,1} +\operatorname{Tr}(\bm{X}^{(v)} \bm{L}_{\bm{G}}{\bm{X}^{(v)}}^T)$. According to~\cite{hou2017multi}, we can obtain the optimal solution of $\alpha^{(v)}$ as follows:
\begin{equation}\label{op-alpha2}
    \begin{aligned}
    {{\alpha} ^{(v)}} = \frac{{(z^{(v)})}^{-1}}{\sum_{v = 1}^l {(z^{(v)})}^{-1}}. 
    \end{aligned}
\end{equation}

The optimization procedure of SHINE-FS is summarized in Algorithm~\ref{code-all}. 
In this algorithm, $\bm{G}$ is initialized as a $k$-nearest neighbor graph constructed from the concatenated data of all views~\cite{wang2019gmc}. $\bm{A}$, $\bm{C}$, and $\bm{S}$ are randomly initialized subject to their respective constraints. Moreover, $\bm{D}^{(v)}$ and $\alpha^{(v)}$ are initialized as the identity matrix and $\frac{1}{l}$, respectively.

\subsection{Time Complexity and Convergence Analysis} 
Algorithm~\ref{code-all} involves the alternating update of seven variables. In each iteration, $\bm{W}^{(v)}$ is updated using the GPI method with a computational complexity of $\mathcal{O}(d_v^2 c)$. Updating $\bm{A}$ costs $\mathcal{O}(n^2)$. Updating $\bm{C}$ takes $\mathcal{O}(d_v c n + c n m + m c^2)$ complexity, where $m \ll d_v$ and $c \ll n$. The computational complexity for updating $\bm{F}$ is $\mathcal{O}(n^2 c)$. Updating $\bm{S}$, $\bm{G}$, and $\alpha^{(v)}$ involves only element-wise operations, so the computational cost is negligible. In summary, the time complexity per iteration of Algorithm~\ref{code-all} is $\mathcal{O}(n^2 c + \sum_{v=1}^l d_v c (d_v + n))$.

Since the objective function in Eq.~\eqref{fun-all} is non-convex with respect to all variables, we optimize it using an alternative iterative optimization algorithm, as described in Algorithm~\ref{code-all}. Here, we provide a brief convergence analysis for each step. 
For step 3 of Algorithm~\ref{code-all}, the convergence of the GPI method used to update $\bm{W}^{(v)}$ is theoretically supported by~\cite{nie2017generalized}. In steps 4, 6, 7, and 9, the closed-form solutions to Eqs.~\eqref{op-a3},~\eqref{op-s3},~\eqref{op-g2}, and~\eqref{op-alpha2} ensure the convergence of updates for $\bm{A}$, $\bm{S}$, $\bm{G}$, and $\alpha^{(v)}$, respectively. For steps 5 and 8, which update $\bm{C}$ and $\bm{F}$, the objective values of Eqs.~\eqref{op-c2} and~\eqref{op-f} are guaranteed to decrease monotonically, as demonstrated in~\cite{nie2016unsupervised}. Furthermore, the convergence behavior of Algorithm~\ref{code-all} will be empirically validated in the experimental section.

\section{Experiments}
In this section, comprehensive experiments are conducted on multiple real-world datasets to evaluate the effectiveness of SHINE-FS in comparison with state-of-the-art approaches.

\subsection{Experimental Schemes}

\begin{table}[t]
    \tabcolsep 0pt 
    \caption{A detailed description of datasets} 
\small \label{data-describe} 
    \vspace*{-10pt} 
    \renewcommand\tabcolsep{1.5pt} 
    \begin{flushleft} 
        \def\temptablewidth{\textwidth} 
        \resizebox{0.485\temptablewidth}{!}{
            \begin{tabular}{@{\extracolsep{\fill}}lccccc}
                \toprule              
                Datasets & Abbr. & \# Instances & \# Classes & \# Views & \# Features \\
                \midrule               
                Yale & Yale & 165 & 15 & 3 & 4096/3304/6750 \\
                MSRA & MSRA & 210 & 7 & 6 & 1302/48/512/100/256/210 \\
                Politics-ie & Poli & 348 & 7 & 9 & \makecell{14377/1047/1051/348/\\348/348/348/348/348} \\
                Mfeat & Mfeat & 2000 & 10 & 6 & 216/76/64/6/240/47 \\
                Scene-15 & Scene & 4485 & 15 & 3 & 20/59/40 \\
                USPS-MNIST & USPS & 9298 & 10 & 2 & 256/32 \\
                Aloi & Aloi & 11025 & 100 & 4 & 77/13/64/64 \\
                SensIT Vehicle & Sens & 15000 & 3 & 2 & 50/50 \\
                \bottomrule
        \end{tabular}}
    \end{flushleft}
\end{table}

\subsubsection{Datasets}
In this study, we use eight real-world multi-view datasets in our experiments, which are described as follows:

Yale~\cite{yale}: 165 grayscale facial images from 15 individuals, with 11 images per subject captured under various lighting conditions and facial expressions.

MSRA~\cite{msra}: 210 object images divided into seven categories, each represented by six distinct views: CENT, CMT, GIST, HOG, LBP, and SIFT.

Politics-ie~\cite{poli}: 348 Twitter users divided into seven groups based on political affiliation, with each user characterized by nine views covering both network and textual information.

Mfeat~\cite{wang2019gmc}: 2,000 handwritten digit samples, each described by six feature sets extracted using different descriptors.

Scene-15~\cite{Scene}: 4,485 scene images across 15 indoor and outdoor categories, with each image represented by three views: GIST, PHOG, and LBP.

USPS-MNIST~\cite{USPS}: 9,298 handwritten digit samples belonging to 10 classes, with each sample obtained from two sources: USPS and MNIST.

Aloi~\cite{ALOI}: 11,025 images of 100 unique objects, photographed under diverse illumination and viewing angles, with each image represented by four feature sets: RGB, HSV, color similarity, and Haralick descriptors.

SensIT Vehicle~\cite{sensit_Cai}: 30,000 instances collected via a wireless sensor network, with each instance capturing signals from acoustic and seismic sensors for vehicle classification. Following~\cite{zhao2020co}, a subset of 15,000 samples is constructed by randomly selecting 5,000 samples from each class.

A detailed description of these multi-view datasets is provided in Table~\ref{data-describe}.

\subsubsection{Evaluation Metrics}
Following~\cite{EMUFS2024, unifier2024}, we adopt two commonly used metrics in MUFS to assess the quality of the selected features, namely Clustering Accuracy (ACC) and Normalized Mutual Information (NMI). Their definitions are as follows:
\begin{equation}\label{ACC}
	\mathrm{ACC} = \frac{1}{n}\sum\limits_{i = 1}^n {\delta(y_i, \mathnormal{map}(y_{i}'))} ,
\end{equation}
where $n$ is the total number of samples, and $y_i$ and $y_{i}'$ correspond to the ground-truth and predicted cluster labels of the $i$-th sample, respectively. The function $\mathnormal{map}(\cdot)$ matches each cluster to its optimal true label using the Kuhn–Munkres algorithm~\cite{KM}, and $\delta(x, y)$ is an indicator function that equals 1 when $x = y$ and 0 otherwise.

\begin{equation}\label{NMI}
	\mathrm{NMI} = \frac{\mathnormal{MI}(\mathcal{C},\mathcal{C'})}{\mathnormal{max} ( H( \mathcal{C}),H(\mathcal{C'}))},
\end{equation}
where $\mathcal{C}$ and $\mathcal{C'}$ denote the sets of ground-truth and predicted clusters, respectively. $\mathnormal{MI}(\mathcal{C}, \mathcal{C'})$  is the mutual information between $\mathcal{C}$ and $\mathcal{C'}$, and $H( \mathcal{C})$ and $H(\mathcal{C'})$ are their corresponding entropies. For both metrics above, higher values indicate better performance.

\begin{table*}[!htbp]
	\centering
	\caption{Means and standard deviations (\%) of ACC  for various methods across different datasets.  The best results are highlighted in bold.}\label{ACC_table}
	\resizebox{\textwidth}{!}{
		\renewcommand\tabcolsep{5pt}
		\begin{tabular}{lcccccccc}
			\toprule[1pt]
			
			\renewcommand{\arraystretch}{0.8}
			\diagbox[width=7em,height=2em]{Methods}{Datasets} & Yale & MSRA & Poli & Mfeat & Scene & USPS & Aloi & Sens \\
			\midrule[1pt]
            
SHINE-FS & $\mathbf{57.01\pm4.69}$ & $\mathbf{92.70\pm1.26}$ & $\mathbf{58.02\pm4.38}$ & $\mathbf{93.11\pm1.71}$ & $\mathbf{41.96\pm0.75}$ & $\mathbf{75.32\pm0.51}$ & $\mathbf{59.45\pm1.84}$ & $\mathbf{62.04\pm0.00}$ \\
AllFea & $49.41\pm4.34 \bullet$ & $51.19\pm3.40 \bullet$ & $40.80\pm1.84 \bullet$ & $50.31\pm1.21 \bullet$ & $29.45\pm0.77 \bullet$ & $35.50\pm0.34 \bullet$ & $16.76\pm0.20 \bullet$ & $60.45\pm0.01 \bullet$ \\
LPScore & $43.70\pm3.47 \bullet$ & $52.43\pm0.46 \bullet$ & $40.57\pm1.59 \bullet$ & $46.93\pm0.16 \bullet$ & $28.68\pm0.22 \bullet$ & $35.46\pm0.24 \bullet$ & $16.71\pm0.23 \bullet$ & $59.85\pm0.01 \bullet$ \\
HSL & $48.42\pm2.83 \bullet$ & $51.14\pm3.27 \bullet$ & $40.37\pm1.34 \bullet$ & $49.04\pm0.82 \bullet$ & $28.84\pm1.03 \bullet$ & $35.48\pm0.22 \bullet$ & $16.72\pm0.20 \bullet$ & $60.00\pm0.00 \bullet$ \\
GAWFS & $50.36\pm3.77 \bullet$ & $62.11\pm2.95 \bullet$ & $41.62\pm2.56 \bullet$ & $41.61\pm0.95 \bullet$ & $25.88\pm0.81 \bullet$ & $24.52\pm0.03 \bullet$ & $16.55\pm0.74 \bullet$ & $44.69\pm0.00 \bullet$ \\
GCDUFS & $56.30\pm4.50 \circ$ & $79.13\pm3.54 \bullet$ & $42.33\pm3.39 \bullet$ & $59.17\pm2.42 \bullet$ & $36.37\pm0.39 \bullet$ & $32.15\pm0.78 \bullet$ & $45.12\pm1.37 \bullet$ & $49.04\pm0.42 \bullet$ \\
UKMFS & $54.51\pm3.68 \bullet$ & $87.48\pm1.19 \bullet$ & $50.76\pm6.29 \bullet$ & $66.89\pm5.66 \bullet$ & $31.79\pm0.57 \bullet$ & $60.75\pm0.24 \bullet$ & $52.96\pm1.42 \bullet$ & $58.49\pm0.00 \bullet$ \\
CFSMO & $45.78\pm3.02 \bullet$ & $51.81\pm2.09 \bullet$ & $40.89\pm1.94 \bullet$ & $67.22\pm0.38 \bullet$ & $28.18\pm0.35 \bullet$ & $35.56\pm0.38 \bullet$ & $46.00\pm1.71 \bullet$ & $59.88\pm0.00 \bullet$ \\
EUMFS & $49.33\pm4.57 \bullet$ & $58.06\pm1.22 \bullet$ & $42.03\pm2.26 \bullet$ & $69.41\pm0.94 \bullet$ & $32.44\pm0.15 \bullet$ & $54.38\pm0.49 \bullet$ & $51.66\pm1.29 \bullet$ & $52.20\pm0.23 \bullet$ \\
CDMvFS & $47.86\pm3.17 \bullet$ & $53.14\pm1.11 \bullet$ & $40.78\pm1.21 \bullet$ & $70.27\pm2.32 \bullet$ & $37.80\pm0.42 \bullet$ & $35.63\pm0.03 \bullet$ & $34.03\pm0.77 \bullet$ & $59.81\pm0.00 \bullet$ \\
JMVFG & $55.74\pm4.83 \circ$ & $78.92\pm1.43 \bullet$ & $40.02\pm3.61 \bullet$ & $72.78\pm0.70 \bullet$ & $35.93\pm0.26 \bullet$ & $58.71\pm0.14 \bullet$ & $46.89\pm1.25 \bullet$ & $47.67\pm0.02 \bullet$ \\
SDFS & $56.97\pm4.29 \circ$ & $31.92\pm1.61 \bullet$ & $41.89\pm7.00 \bullet$ & $74.21\pm4.00 \bullet$ & $36.60\pm0.92 \bullet$ & $59.96\pm0.13 \bullet$ & $48.76\pm1.50 \bullet$ & $60.00\pm0.00 \bullet$ \\
CCSFS & $47.88\pm3.48 \bullet$ & $62.94\pm5.35 \bullet$ & $40.80\pm1.85 \bullet$ & $46.20\pm1.14 \bullet$ & $27.66\pm0.49 \bullet$ & $36.24\pm0.10 \bullet$ & $53.19\pm1.86 \bullet$ & $60.00\pm0.00 \bullet$ \\
TLR-MUFS & $47.92\pm4.49 \bullet$ & $51.22\pm3.52 \bullet$ & $40.95\pm1.21 \bullet$ & $39.13\pm0.24 \bullet$ & $34.95\pm0.48 \bullet$ & $35.59\pm0.04 \bullet$ & $16.68\pm0.28 \bullet$ & $43.55\pm0.00 \bullet$ \\

			\bottomrule[1pt]
		\end{tabular}
	}
\end{table*}

\begin{table*}[!htbp]
	\centering
	\caption{Means and standard deviations (\%) of NMI for various methods across different datasets.  The best results are highlighted in bold.}\label{NMI_table}
	\resizebox{\textwidth}{!}{
		\renewcommand\tabcolsep{5pt}
		\begin{tabular}{lcccccccc}
			\toprule[1pt]
			
			\renewcommand{\arraystretch}{0.8}
			\diagbox[width=7em,height=2em]{Methods}{Datasets} & Yale & MSRA & Poli & Mfeat & Scene & USPS & Aloi & Sens \\
			\midrule[1pt]
            SHINE-FS & $\mathbf{62.92\pm3.02}$ & $\mathbf{86.48\pm1.73}$ & $\mathbf{54.71\pm3.12}$ & $\mathbf{86.62\pm0.97}$ & $\mathbf{40.49\pm0.44}$ & $\mathbf{60.23\pm0.33}$ & $\mathbf{77.45\pm0.73}$ & $\mathbf{18.59\pm0.00}$ \\
AllFea & $55.48\pm2.64 \bullet$ & $41.16\pm0.86 \bullet$ & $7.57\pm2.18 \bullet$ & $56.80\pm0.87 \bullet$ & $28.72\pm0.30 \bullet$ & $25.96\pm0.18 \bullet$ & $37.33\pm0.12 \bullet$ & $16.74\pm0.01 \bullet$ \\
LPScore & $50.70\pm2.21 \bullet$ & $41.38\pm0.40 \bullet$ & $8.12\pm2.77 \bullet$ & $53.72\pm0.11 \bullet$ & $25.54\pm0.16 \bullet$ & $25.97\pm0.14 \bullet$ & $37.31\pm0.12 \bullet$ & $15.97\pm0.01 \bullet$ \\
HSL & $53.82\pm1.89 \bullet$ & $40.99\pm1.22 \bullet$ & $6.69\pm1.99 \bullet$ & $55.90\pm0.64 \bullet$ & $27.03\pm0.24 \bullet$ & $25.97\pm0.14 \bullet$ & $37.29\pm0.14 \bullet$ & $16.18\pm0.00 \bullet$ \\
GAWFS & $55.75\pm2.72 \bullet$ & $51.00\pm2.27 \bullet$ & $12.69\pm3.14 \bullet$ & $29.53\pm0.61 \bullet$ & $25.50\pm0.41 \bullet$ & $14.30\pm0.10 \bullet$ & $32.27\pm1.27 \bullet$ & $7.11\pm0.00 \bullet$ \\
GCDUFS & $61.03\pm3.66 \bullet$ & $70.02\pm2.06 \bullet$ & $31.24\pm3.68 \bullet$ & $63.17\pm1.35 \bullet$ & $37.95\pm0.40 \bullet$ & $23.57\pm0.63 \bullet$ & $66.31\pm0.75 \bullet$ & $9.38\pm0.41 \bullet$ \\
UKMFS & $60.64\pm3.48 \bullet$ & $79.14\pm1.72 \bullet$ & $48.15\pm5.08 \bullet$ & $71.08\pm3.41 \bullet$ & $34.59\pm0.36 \bullet$ & $51.24\pm0.17 \bullet$ & $73.67\pm0.62 \bullet$ & $14.21\pm0.00 \bullet$ \\
CFSMO & $52.00\pm2.18 \bullet$ & $41.12\pm0.47 \bullet$ & $7.93\pm2.82 \bullet$ & $62.39\pm0.36 \bullet$ & $26.05\pm0.18 \bullet$ & $25.92\pm0.21 \bullet$ & $69.78\pm0.71 \bullet$ & $16.04\pm0.00 \bullet$ \\
EUMFS & $54.67\pm3.61 \bullet$ & $47.96\pm1.17 \bullet$ & $11.21\pm3.45 \bullet$ & $66.51\pm0.74 \bullet$ & $30.22\pm0.11 \bullet$ & $51.76\pm0.14 \bullet$ & $72.02\pm0.53 \bullet$ & $9.93\pm0.09 \bullet$ \\
CDMvFS & $53.23\pm2.56 \bullet$ & $43.42\pm1.21 \bullet$ & $6.89\pm2.26 \bullet$ & $66.79\pm1.61 \bullet$ & $36.16\pm0.51 \bullet$ & $26.48\pm0.02 \bullet$ & $56.58\pm0.37 \bullet$ & $15.94\pm0.00 \bullet$ \\
JMVFG & $61.08\pm4.37 \bullet$ & $69.74\pm1.71 \bullet$ & $8.94\pm3.00 \bullet$ & $63.95\pm0.65 \bullet$ & $37.65\pm0.25 \bullet$ & $52.92\pm0.08 \bullet$ & $66.43\pm0.55 \bullet$ & $8.82\pm0.02 \bullet$ \\
SDFS & $60.41\pm4.27 \bullet$ & $17.01\pm2.06 \bullet$ & $15.01\pm4.58 \bullet$ & $72.17\pm1.71 \bullet$ & $34.46\pm0.59 \bullet$ & $51.60\pm0.08 \bullet$ & $70.46\pm0.72 \bullet$ & $16.17\pm0.00 \bullet$ \\
CCSFS & $53.53\pm2.55 \bullet$ & $55.90\pm2.49 \bullet$ & $7.93\pm2.72 \bullet$ & $41.70\pm0.42 \bullet$ & $28.08\pm0.41 \bullet$ & $26.51\pm0.02 \bullet$ & $74.44\pm0.69 \bullet$ & $16.17\pm0.00 \bullet$ \\
TLR-MUFS & $53.37\pm3.28 \bullet$ & $45.21\pm2.12 \bullet$ & $7.40\pm2.05 \bullet$ & $46.68\pm0.10 \bullet$ & $38.17\pm0.32 \bullet$ & $25.89\pm0.01 \bullet$ & $37.29\pm0.14 \bullet$ & $5.14\pm0.00 \bullet$ \\

			\bottomrule[1pt]
		\end{tabular}
	}
\end{table*}

\begin{figure*}[!t]
  \centering
    \makebox[\textwidth][c]{\includegraphics[width=0.952\textwidth]{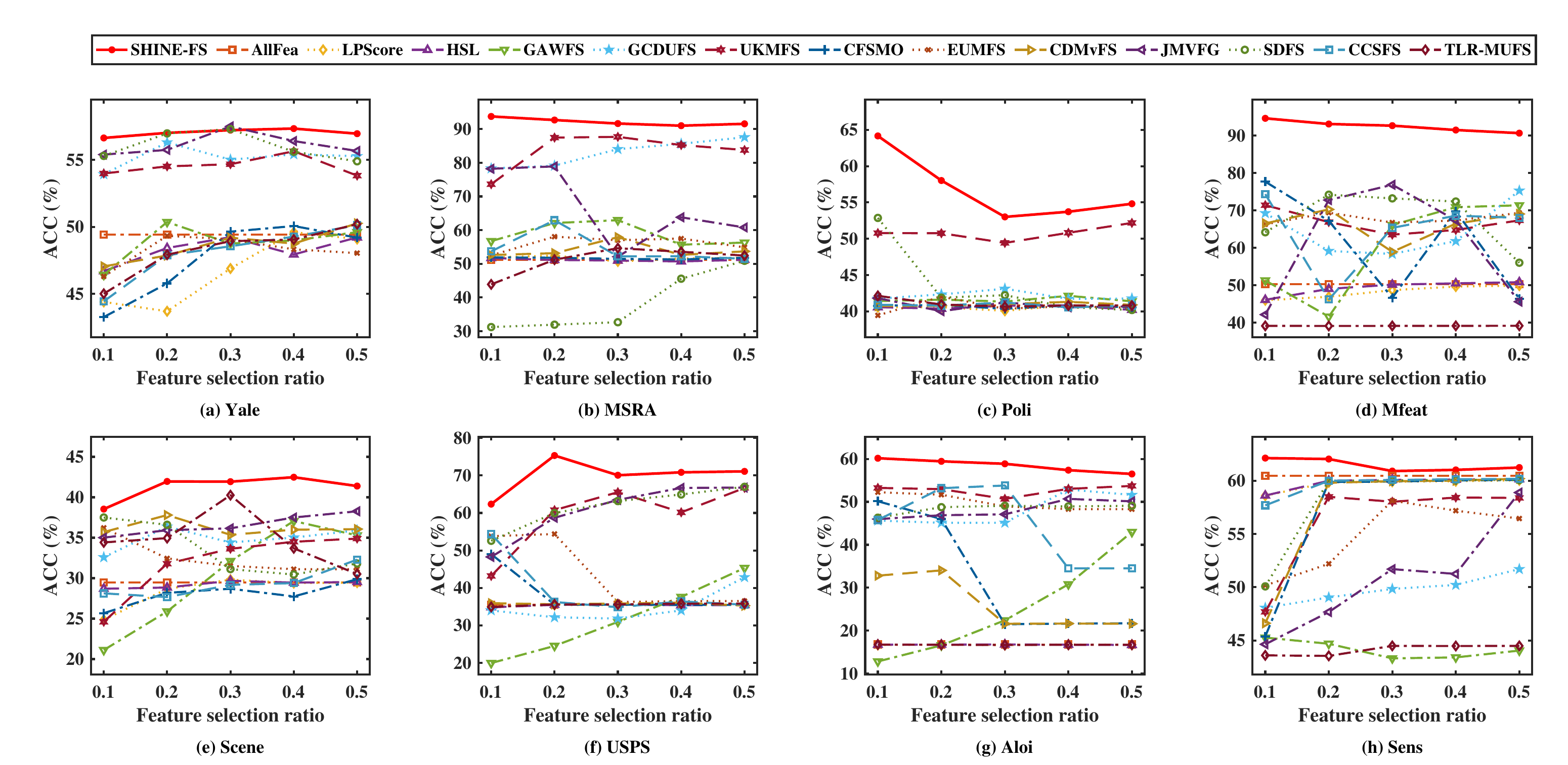}}
    \caption{ACC of different methods on eight datasets with different feature selection ratios.}
    \label{ACC-8}

  \vspace{\floatsep}

    \makebox[\textwidth][c]{\includegraphics[width=0.952\textwidth]{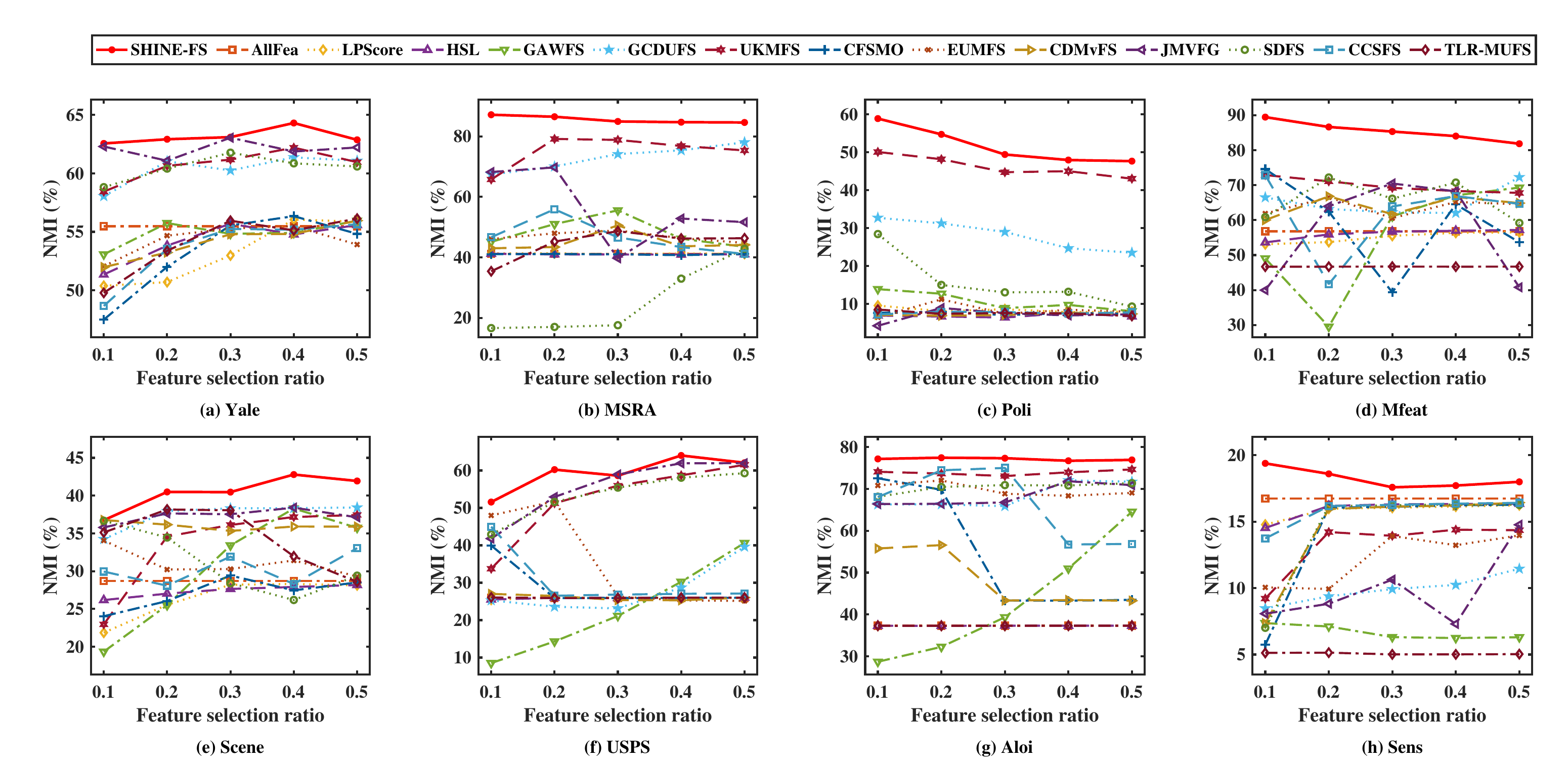}}
    \caption{NMI of different methods on eight datasets with different feature selection ratios.}
    \label{NMI-8}
\end{figure*}

\subsubsection{Compared Methods}
To evaluate the effectiveness of SHINE-FS, we compare it with several state-of-the-art (SOTA) methods, including three single-view methods (LPScore, HSL, and GAWFS) and nine multi-view methods (GCDUFS, UKMFS, CFSMO, EMUFS, CDMvFS, JMVFG, SDFS, CCSFS, and TLR-MUFS). Below is a brief introduction to these methods.

\begin{itemize}
	\item \textbf{AllFea} utilizes all features from different views for comparison. 
	\item \textbf{LPScore}~\cite{LS2005}  performs feature selection using the Laplacian score, which assesses the importance of each feature based on its ability to preserve local data structure.
	\item \textbf{HSL}~\cite{HSL2024} integrates feature selection with the learning of both first-order and higher-order similarities within a unified framework.
	\item \textbf{GAWFS}~\cite{GAWFS2025} simultaneously performs clustering and feature selection by combining non-negative matrix factorization with adaptive graph learning.
	\item \textbf{GCDUFS}~\cite{GCDUFS2025}  identifies important features by leveraging both consensus information across views and inter-view diversity.
	\item \textbf{UKMFS}~\cite{UKMFS2025} proposes a unified feature selection framework that incorporates robust self-representation and binary hashing.
	\item \textbf{CFSMO}~\cite{CFSMO2024} utilizes multi-order neighborhood information from similarity graphs, together with consistency across views, to enhance feature selection.
	\item \textbf{EMUFS}~\cite{EMUFS2024}  jointly learns cluster assignments and local data structures by integrating membership matrices and bipartite graphs to identify discriminative features.
	\item \textbf{CDMvFS}~\cite{CDMvFS2024}	combines similarity graph learning with spectral clustering to select informative features.
	\item \textbf{JMVFG}~\cite{JMVFG2024} integrates orthogonal decomposition and consensus graph learning to preserve local data structure for feature selection.
	\item \textbf{SDFS}~\cite{SDFS2023} simultaneously learns view-specific and consensus similarity graphs using an automatic weighting strategy to preserve data locality and guide feature selection.
	\item \textbf{CCSFS}~\cite{CCSFS2023} unifies subspace learning, consensus cluster discovery, and feature selection within a single framework.
	\item \textbf{TLR-MUFS}~\cite{TLR2022}  identifies important features by capturing intra-view structures from view-specific graphs and enforcing consistency across views via tensor low-rank regularization.
\end{itemize}

 To ensure a fair comparison, we use grid search to tune the parameters of all methods and report their best performance. In our method, the parameters $\gamma$, $\beta$, and $\eta$ are tuned using the same strategy within the range $\{10^{-3}, 10^{-2}, 10^{-1}, 1, 10^{1}, 10^{2}, 10^{3}\}$. Since selecting the optimal number of features for a given dataset remains challenging~\cite{li2017feature, Cao2024Partition}, we vary the percentage of selected features from 10\% to 50\% in increments of 10\% for all datasets. Then, k-means clustering~\cite{kmeans} is performed 30 times on the selected features, and the average results are reported. All experiments were carried out using MATLAB R2022b on a desktop with an Intel Core i9-10900 CPU (2.80 GHz) and 64 GB of RAM.

\subsection{Performance Comparison}
In this section, we assess the clustering performance of SHINE-FS in comparison with other methods using ACC and NMI metrics. Tables~\ref{ACC_table} and~\ref{NMI_table} show the ACC and NMI results on eight datasets with a feature selection ratio of 20\%. To further demonstrate the superiority of SHINE-FS, we use the Wilcoxon rank-sum test~\cite{woolson2005wilcoxon} to determine whether its performance is significantly better than that of the competing approaches. In Tables~\ref{ACC_table} and~\ref{NMI_table}, values marked with $\bullet$ indicate that SHINE-FS performs significantly better than the compared methods at the 0.05 significance level, while values marked with $\circ$ indicate that SHINE-FS does not show a significant improvement over the compared methods.

\begin{figure*}[t]
    \centering
    \includegraphics[width=2.05\columnwidth]{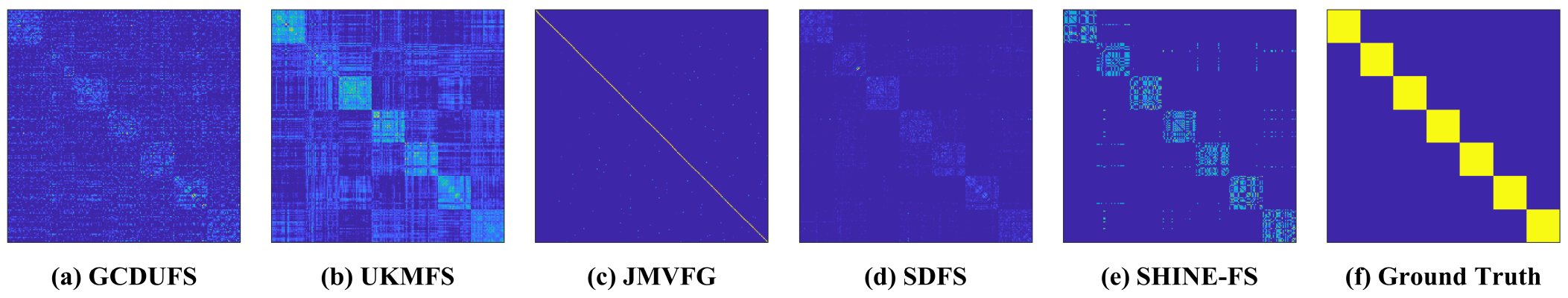} 
    \caption{Visualizations of the similarity graphs learned by different graph-based methods on MSRA dataset.}
    \label{graph}
\end{figure*}

\begin{figure}[t]
    \centering
    \includegraphics[width=0.98\columnwidth]{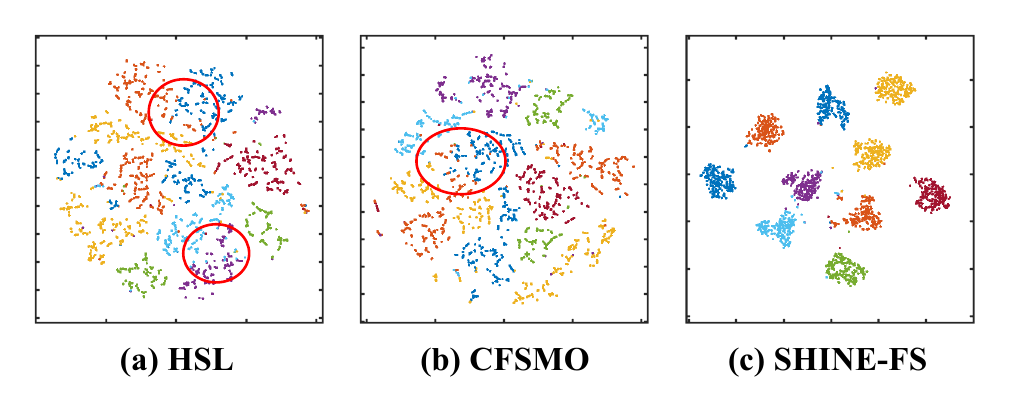} 
    \caption{The t-SNE visualizations on Mfeat dataset.}
    \label{tsne}
\end{figure}

As shown in Tables~\ref{ACC_table} and~\ref{NMI_table}, the proposed method consistently outperforms the compared approaches. As to MSRA, Mfeat, and USPS datasets, SHINE-FS achieves average improvements of more than 34\% in ACC and 31\% in NMI. For Poli and Aloi datasets, SHINE-FS achieves average improvements exceeding 20\% in ACC and 31\% in NMI. For Yale, Scene, and Sens datasets, SHINE-FS also outperforms the competing methods, with average increases of over 7\% in both ACC and NMI.

Furthermore, since it is difficult to determine the optimal number of selected features, we compare the performance of various methods under different feature selection ratios in terms of ACC and NMI. The ACC and NMI results are illustrated in Figs.~\ref{ACC-8} and~\ref{NMI-8}, respectively. As observed in these figures, the proposed method achieves superior performance over the competing methods in most cases. The superior performance of SHINE-FS can be attributed to its unified framework, which seamlessly combines multi-view feature selection, consensus anchor graph learning, and hybrid-order similarity learning. This integration enables the model to capture both local and global structures, providing a more effective characterization of the intrinsic data structure, thus better guiding feature selection.

\subsection{Visualization Analysis}
To intuitively demonstrate the effectiveness of the learned hybrid-order similarity graph in capturing the intrinsic data structure, we visualize the similarity graphs generated by different graph-based methods on MSRA dataset, as shown in Fig.~\ref{graph}. It can be observed that the hybrid-order similarity graph learned by SHINE-FS (Fig.~\ref{graph}~(e)) exhibits a clearer structure that is highly consistent with the ground-truth graph (Fig.~\ref{graph}~(f)). In contrast, the graphs generated by other methods (Figs.~\ref{graph}~(a)-(d)) show ambiguous structures that poorly represent the sample relationships. These results demonstrate the superiority of SHINE-FS in uncovering the intrinsic data structure.

Furthermore, to visually evaluate the quality of the selected features, we use t-SNE~\cite{maaten2008visualizing} to embed the samples represented by these features into a two-dimensional space. Fig.~\ref{tsne} illustrates the t-SNE visualization results on Mfeat dataset with a 20\% feature selection ratio.  Figs.~\ref{tsne}~(a) and (b) show the results of HSL and CFSMO, respectively, both of which rely on predefined high-order similarities, while Fig.~\ref{tsne}~(c) presents the visualization result of SHINE-FS. The results indicate that HSL and CFSMO show considerable overlap among clusters, while SHINE-FS produces greater inter-cluster separation and clearer boundaries. This demonstrates that SHINE-FS can effectively select a compact subset of discriminative features by adaptively learning a hybrid-order similarity graph.

\begin{figure}[t!] 
    \centering
    \includegraphics[width=1\columnwidth]{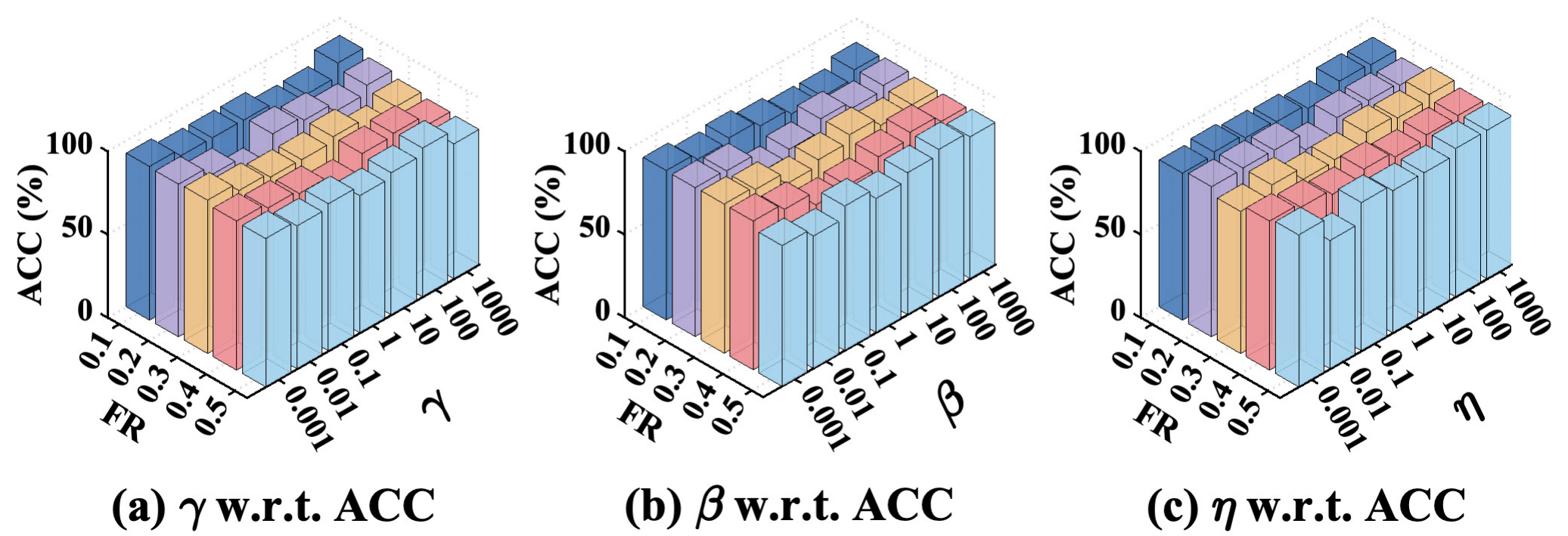} 
    \caption{ACC of SHINE-FS with varying parameters $\gamma$, $\beta$, $\eta$ and feature selection ratios on MSRA dataset.}
    \label{Sensi-ACC}
    \vspace{\floatsep} 
    \includegraphics[width=1\columnwidth]{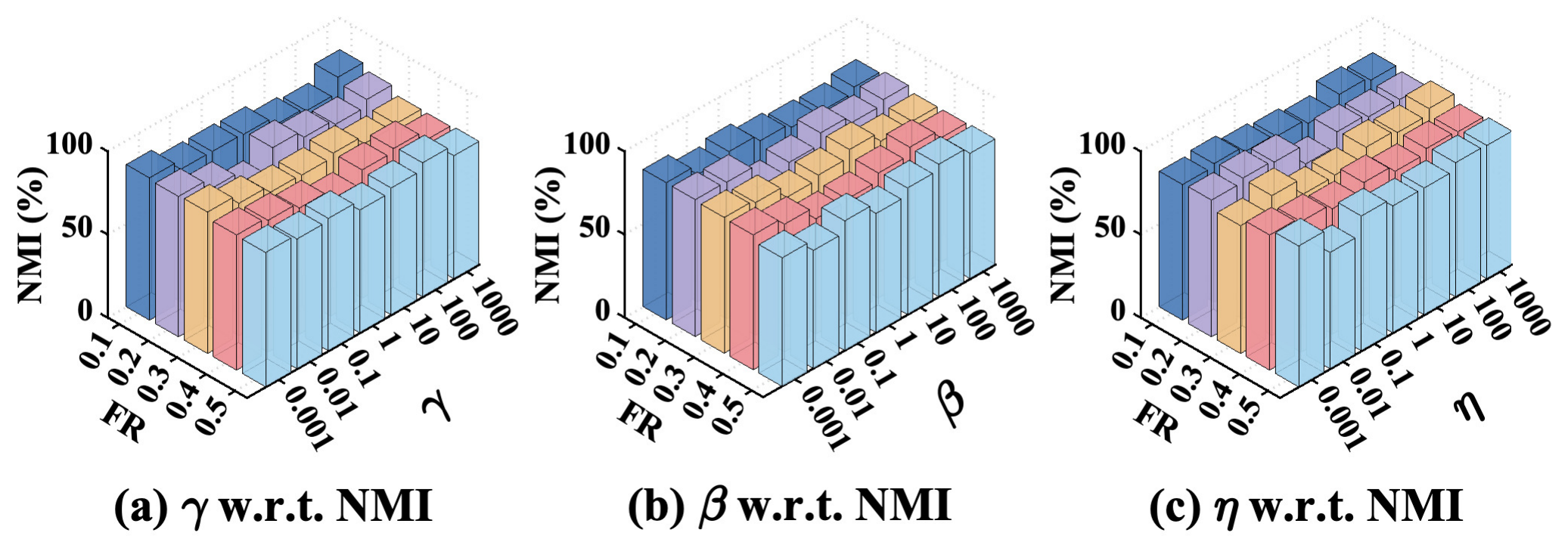} 
    \caption{NMI of SHINE-FS with varying parameters $\gamma$, $\beta$, $\eta$ and feature selection ratios on MSRA dataset.}
    \label{Sensi-NMI}
    \vspace{\floatsep}
    \includegraphics[width=1.01\columnwidth]{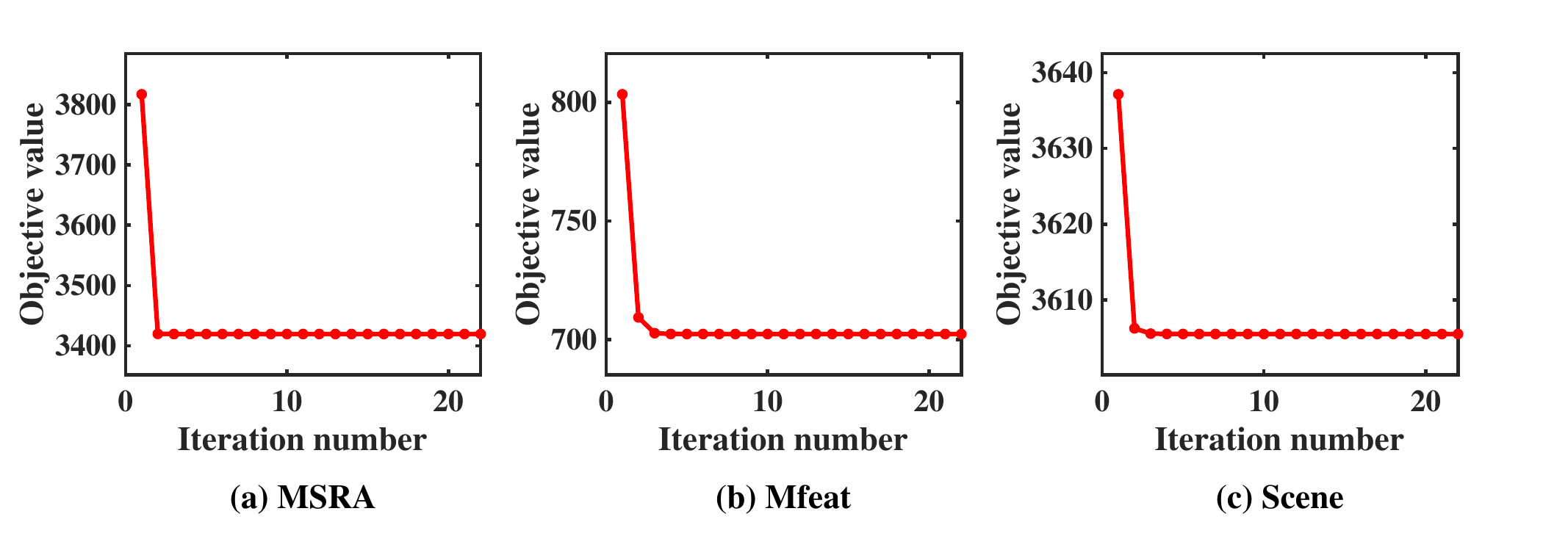} 
    \caption{Convergence curves of SHINE-FS on MSRA, Mfeat and Scene datasets.}
    \label{Conver}
\end{figure}

\subsection{Parameter Sensitivity and Convergence Behavior}
The objective function in Eq.~\eqref{fun-all} includes five parameters: $\gamma$, $\beta$, $\eta$, $\lambda_{S}$, and $\lambda_{G}$. Among these, $\lambda_{S}$ and $\lambda_{G}$ are automatically determined when solving for $\bm{S}$ in Eq.~\eqref{op-s3} and $\bm{G}$ in Eq.~\eqref{op-g2}, respectively. 
To investigate the sensitivity of SHINE-FS to the parameters $\gamma$, $\beta$, and $\eta$, Figs.~\ref{Sensi-ACC} and~\ref{Sensi-NMI} illustrate how its ACC and NMI performance changes as these parameters and the feature selection ratio (FR) are varied on the MSRA dataset. As illustrated in these figures, SHINE-FS remains relatively stable for $\gamma$ and $\beta$, while slight fluctuations are observed as $\eta$ varies.  In practice, the parameter $\eta$ should be tuned using a grid search to achieve optimal performance. 

Furthermore, to verify the convergence of Algorithm 1, we present its convergence behavior curves on MSRA, Mfeat, and Scene datasets. As shown in Fig.~\ref{Conver}, the objective value of SHINE-FS drops sharply during the initial iterations and stabilizes rapidly, typically within the first 10 iterations. This demonstrates that our method converges quickly.

\subsection{Ablation Study}
\begin{figure*}[!htbp]
    \centering  \includegraphics[width=1.95\columnwidth]{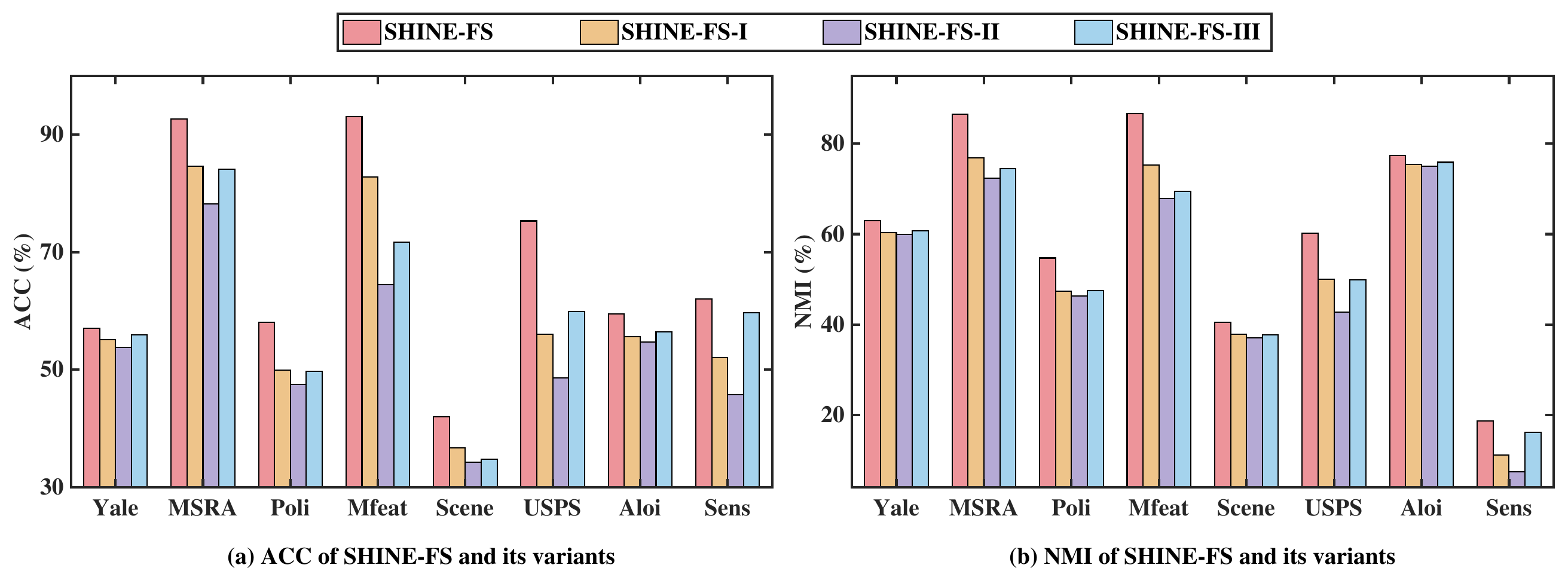} 
    \caption{Performance comparison of SHINE-FS and its three variants in terms of ACC and NMI.}
    \label{ablation}
\end{figure*}

To demonstrate the effectiveness of the proposed module in SHINE-FS, we conduct ablation experiments by comparing it with three of its variants.
Specifically, SHINE-FS-I excludes only the anchor-induced second-order similarity graph learning module. SHINE-FS-II removes the hybrid-order similarity graph learning module. SHINE-FS-III replaces the consensus anchor graph learning module with a view-specific anchor graph learning module. Fig.~\ref{ablation} presents the ablation experiment results in terms of ACC and NMI on eight datasets. It can be observed that: (\romannumeral1) SHINE-FS consistently outperforms SHINE-FS-I, validating that the anchor-induced second-order similarity graph can effectively capture the global structure of the data.
(\romannumeral2) The superior performance of SHINE-FS over SHINE-FS-II demonstrates the effectiveness of adaptive hybrid-order similarity graph learning for jointly capturing local and global structures, thus enabling a more comprehensive characterization of the intrinsic data structure.
(\romannumeral3)  SHINE-FS outperforms SHINE-FS-III, indicating that consensus anchor graph learning exploits shared structural information across views to facilitate feature selection.

\section{Conclusion}
In this paper, we propose a novel MUFS method called SHINE-FS to address the challenge of capturing the intrinsic data structure in MUFS. SHINE-FS first learns consensus anchors and the corresponding anchor graph to capture anchor-sample relationships across views, which facilitates the learning of a second-order similarity graph. Furthermore, SHINE-FS jointly learns the first-order and second-order similarity graphs to construct a hybrid similarity graph that effectively integrates local and global structural information, thus better characterizing the intrinsic data structure.  Extensive experiments on real multi-view datasets demonstrate the superiority of SHINE-FS over SOTA methods. In many practical applications, labeled samples are often scarce; however, they provide essential discriminative information that can effectively guide feature selection. Therefore, our future work will aim to extend the proposed unsupervised framework to a semi-supervised scenario, enabling the identification of informative features from multi-view data with limited labels available.

\bibliographystyle{IEEEtran}
\bibliography{ref}

\vfill

\end{document}